%% file: paper.tex
\documentclass[10pt,twocolumn,letterpaper]{article}

\usepackage{iccv}
\usepackage{times}
\usepackage{epsfig}
\usepackage{graphicx}
\usepackage{amsmath}
\usepackage{amssymb}

\usepackage{color}
\usepackage{listings}
\definecolor{codegreen}{rgb}{0,0.5,0}
\definecolor{codeblue}{rgb}{0.25,0.5,0.5}
\definecolor{codegray}{rgb}{0.6,0.6,0.6}

\newcommand{\mha}{\textsf{MHA}}
\newcommand{\concat}{\textsf{Concat}}
\newcommand{\conv}{\textsf{Conv}}
\newcommand{\softmax}{\textsf{Softmax}}
\newcommand{\aaconv}{\textsf{AAConv}}

\usepackage[pagebackref=true,breaklinks=true,letterpaper=true,colorlinks,bookmarks=false]{hyperref}

\iccvfinalcopy 

\def\iccvPaperID{6563} 

\ificcvfinal\fi
\begin{document}

\title{Attention Augmented Convolutional Networks}

\author{Irwan Bello\\
\and
Barret Zoph\\
\and
Ashish Vaswani\\
\and 
Jonathon Shlens\\
\and
Quoc V. Le\\
Google Brain\\
{\tt\small \{ibello,barretzoph,avaswani,shlens,qvl\}@google.com}
}

\maketitle
\ificcvfinal\thispagestyle{empty}\fi

\input{abstract}
\input{intro}

\input{related}
\input{methods}
\input{results}

\input{discussion}

\section*{Acknowledgements}

The authors would like to thank Tsung-Yi Lin, Prajit Ramachandran, Mingxing Tan, Yanping Huang and the Google Brain team for insightful comments and discussions.

{\small
\bibliographystyle{ieee_fullname}
\bibliography{paper}
}

\input{appendix}

\end{document}

%% file: abstract.tex
\begin{abstract}

Convolutional networks have been the paradigm of choice in many computer vision applications.
The convolution operation however has a significant weakness in that it only operates on a local neighborhood, thus missing global information.
Self-attention, on the other hand, has emerged as a recent advance to capture long range interactions, but has mostly been applied to sequence modeling and generative modeling tasks.
In this paper, we consider the use of self-attention for discriminative visual tasks as an alternative to convolutions.
We introduce a novel two-dimensional relative self-attention mechanism that proves competitive in replacing convolutions as a stand-alone computational primitive for image classification.
We find in control experiments that the best results are obtained when combining both convolutions and self-attention.
We therefore propose to augment convolutional operators with this self-attention mechanism by concatenating convolutional feature maps with a set of feature maps produced via self-attention.
Extensive experiments show that Attention Augmentation leads to consistent improvements in image classification on ImageNet and object detection on COCO across many different models and scales, including ResNets and a state-of-the art mobile constrained network, while keeping the number of parameters similar.
In particular, our method achieves a $1.3\%$ top-1 accuracy improvement on ImageNet classification over a ResNet50 baseline and outperforms other attention mechanisms for images such as Squeeze-and-Excitation~\cite{hu2017squeeze}. 
It also achieves an improvement of 1.4 mAP in COCO Object Detection on top of a RetinaNet baseline.

\end{abstract}

%% file: intro.tex
\section{Introduction}

Convolutional Neural Networks have enjoyed tremendous success in many computer vision applications, especially in image classification~\cite{lecun1998gradient,krizhevsky2012imagenet}. The design of the convolutional layer imposes 1) locality via a limited receptive field and 2) translation equivariance via weight sharing. Both these properties prove to be crucial inductive biases when designing models that operate over images.
However, the local nature of the convolutional kernel prevents it from capturing global contexts in an image, often necessary for better recognition of objects in images~\cite{rabinovich2007objects}.   

\begin{figure}[t]
\centering
\includegraphics[width=\linewidth]{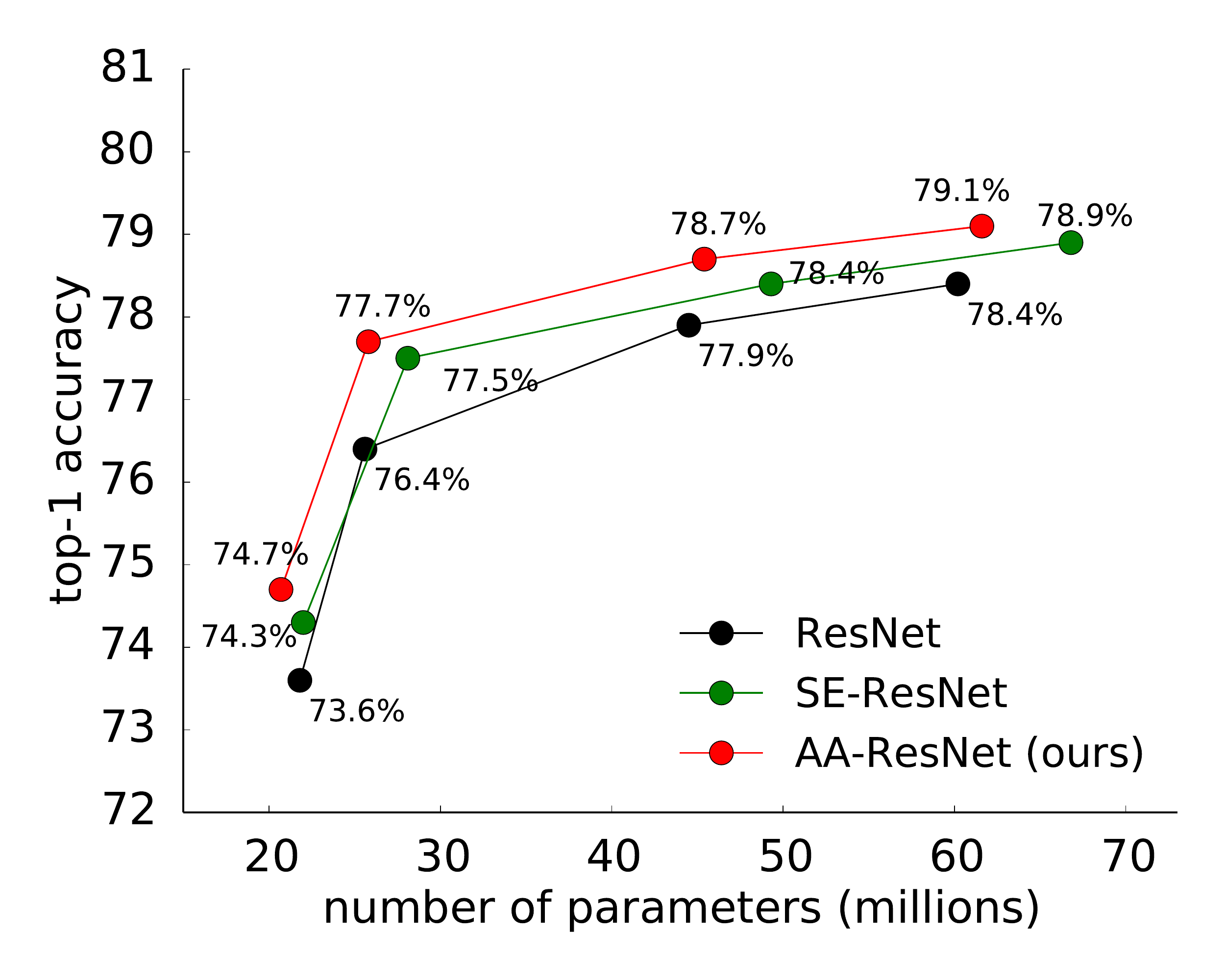}
\caption{\textbf{Attention Augmentation systematically improves image classification across a large variety of networks of different scales.} ImageNet classification accuracy \cite{deng2009imagenet} versus the number of parameters for baseline models (ResNet) \cite{identity-mappings}, models augmented with channel-wise attention (SE-ResNet)~\cite{hu2017squeeze} and our proposed architecture (AA-ResNet).
}
\label{fig:marketing}
\end{figure}

\begin{figure*}[t!]
    \begin{center}
    \includegraphics[width=\textwidth]{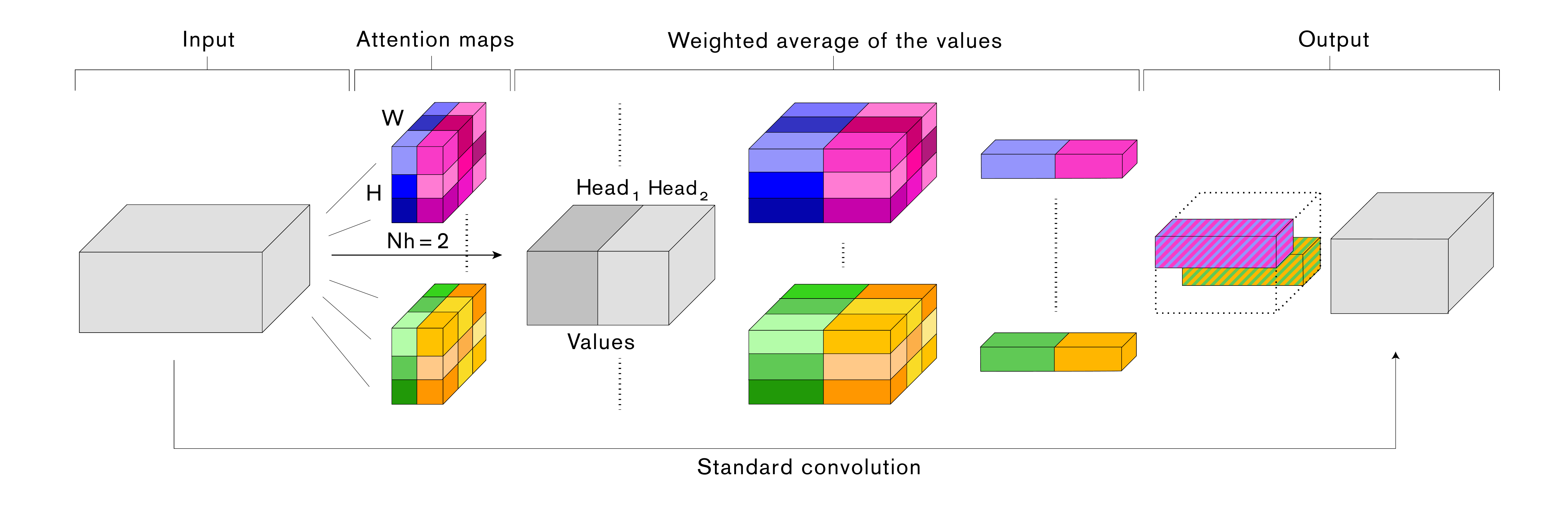}
    \end{center}
    \caption{\textbf{Attention-augmented convolution}:  For each spatial location (h, w), $N_h$ attention maps over the image are computed from queries and keys. These attention maps are used to compute $N_h$ weighted averages of the values V. The results are then concatenated, reshaped to match the original volume's spatial dimensions and mixed with a pointwise convolution. Multi-head attention is applied in parallel to a standard convolution operation and the outputs are concatenated.}
    \label{fig:attention-diagram}
\end{figure*}

Self-attention \cite{vaswani2017attention}, on the other hand, has emerged as a recent advance to capture long range interactions, but has mostly been applied to sequence modeling and generative modeling tasks.
The key idea behind self-attention is to produce a weighted average of values computed from hidden units. 
Unlike the pooling or the convolutional operator, the weights used in the weighted average operation are produced dynamically via a similarity function between hidden units. 
As a result, the interaction between input signals depends on the signals themselves rather than being predetermined by their relative location like in convolutions.
In particular, this allows self-attention to capture long range interactions without increasing the number of parameters.

In this paper, we consider the use of self-attention for discriminative visual tasks as an alternative to convolutions. 
We develop a novel two-dimensional relative self-attention mechanism~\cite{shaw2018self} that maintains translation equivariance while being infused with relative position information, making it well suited for images. 
Our self-attention formulation proves competitive for replacing convolutions entirely, however we find in control experiments that the best results are obtained when combining both.
We therefore do not completely abandon the idea of convolutions, but instead propose to augment convolutions with this self-attention mechanism. 
This is achieved by concatenating convolutional feature maps, which enforce locality, to self-attentional feature maps capable of modeling longer range dependencies (see Figure~\ref{fig:attention-diagram}).

We test our method on the CIFAR-100 and ImageNet classification \cite{krizhevsky2009learning,deng2009imagenet} and the COCO object detection \cite{lin2014microsoft} tasks, across a wide range of architectures at different computational budgets, including a state-of-the art resource constrained architecture~\cite{tan2018mnasnet}. 
Attention Augmentation yields systematic improvements with minimal additional computational burden and notably outperforms the popular Squeeze-and-Excitation~\cite{hu2017squeeze} channelwise attention approach in all experiments. 
In particular, Attention Augmentation achieves a 1.3\% top-1 accuracy ImageNet on top of a ResNet50 baseline and 1.4 mAP increase in COCO object detection on top of a RetinaNet baseline. 
Suprisingly, experiments also reveal that fully self-attentional models, a special case of Attention Augmentation, only perform slightly worse than their fully convolutional counterparts on ImageNet, indicating that self-attention is a powerful stand-alone computational primitive for image classification.

%% file: related.tex
\section{Related Work}

\subsection{Convolutional networks}

Modern computer vision has been built on powerful image featurizers learned on image classification tasks such as CIFAR-10 \cite{krizhevsky2009learning} and ImageNet \cite{deng2009imagenet}. 
These datasets have been used as benchmarks for delineating better image featurizations and network architectures across a broad range of tasks \cite{kornblith2019better}. 
For example, improving the ``backbone'' network typically leads to improvements in object detection \cite{huang2016speed} and image segmentation \cite{chen2018searching}. 
These observations have inspired the research and design of new architectures, which
are typically derived from the composition of convolution operations across an array of spatial scales and skip connections \cite{krizhevsky2012imagenet,szegedy2016rethinking, szegedy2016inception, szegedy2015going,identity-mappings,xie2016aggregated, he2015deep}. 
Indeed, automated search strategies for designing architectures based on convolutional primitives result in state-of-the-art accuracy on large-scale image classification tasks that translate across a range of tasks~\cite{zoph2018learning, kornblith2019better}.

\subsection{Attention mechanisms in networks}

Attention has enjoyed widespread adoption as a computational module for modeling sequences because of its ability to capture long distance interactions~\cite{bahdanau2014neural,vinyals2015ptrnet,bello2016nco,bello2018seq2slate}. 
Most notably, Bahdanau et al.~\cite{bahdanau2014neural} first proposed to combine attention with a Recurrent Neural Network~\cite{lstm} for alignment in Machine Translation.
Attention was further extended by Vaswani et al.~\cite{vaswani2017attention}, where the self-attentional Transformer architecture achieved state-of-the-art results in Machine Translation.
Using self-attention in cooperation with convolutions is a theme shared by recent work in Natural Language Processing~\cite{yang2018csan} and Reinforcement Learning~\cite{santoro2019drlrib}. 
For example, the QANet~\cite{yu2018qanet} and Evolved Transformer~\cite{So2019} architectures alternate between self-attention layers and convolution layers for Question Answering applications and Machine Translation respectively.
Additionally, multiple attention mechanisms have been proposed for visual tasks to address the weaknesses of convolutions~\cite{hu2017squeeze,hu2018gather,chen2018double,woo2018cbam,wang2018non,han18sagan}.
For instance, Squeeze-and-Excitation~\cite{hu2017squeeze} and Gather-Excite~\cite{hu2018gather} reweigh feature channels using signals aggregated from entire feature maps, while BAM~\cite{park2018bam} and CBAM~\cite{woo2018cbam} refine convolutional features \emph{independently} in the channel and spatial dimensions.
In non-local neural networks~\cite{wang2018non}, improvements are shown in video classification and object detection via the additive use of a few non-local residual blocks that employ self-attention in convolutional architectures.
However, non-local blocks are only added to the architecture after ImageNet pretraining and are initialized in such a way that they do not break pretraining.

In contrast, our attention augmented networks do not rely on pretraining of their fully convolutional counterparts and employ self-attention along the entire architecture. 
The use of multi-head attention allows the model to attend \emph{jointly} to both spatial and feature subspaces. 
Additionally, we enhance the representational power of self-attention over images by extending relative self-attention~\cite{shaw2018self,huang2018improved} to two dimensional inputs allowing us to model translation equivariance in a principled way.
Finally our method produces additional feature maps, rather than recalibrating convolutional features via addition~\cite{wang2018non,han18sagan} or gating~\cite{hu2017squeeze,hu2018gather,park2018bam,woo2018cbam}. 
This property allows us to flexibly adjust the fraction of attentional channels and consider a spectrum of architectures, ranging from fully convolutional to fully attentional models.

%% file: methods.tex
\section{Methods}

We now formally describe our proposed Attention Augmentation method.
We use the following naming conventions: $H$, $W$ and $F_{in}$ refer to the height, width and number of input filters of an activation map. $N_h$, $d_v$ and $d_k$ respectively refer the number of heads, the depth of values and the depth of queries and keys in multihead-attention (MHA). We further assume that $N_h$ divides $d_v$ and $d_k$ evenly and denote $d_v^h$ and $d_k^h$ the depth of values and queries/keys per attention head.

\subsection{Self-attention over images}

Given an input tensor of shape $(H, W, F_{in})$,\footnote{We omit the batch dimension for simplicity.} we flatten it to a matrix $X \in \mathbb{R}^{HW \times F_{in}}$ and perform multihead attention as proposed in the Transformer architecture~\cite{vaswani2017attention}. The output of the self-attention mechanism for a single head $h$ can be formulated as:
\begin{equation}
O_h = \softmax\left(\frac{(XW_q)(XW_k)^T}{\sqrt{d_k^h}}\right) (XW_v)
\label{eq:dot-prod-att}
\end{equation}
where $W_q$, $W_k \in \mathbb{R}^{F_{in} \times d_k^h}$ and $W_v \in \mathbb{R}^{F_{in} \times d_v^h}$ are learned linear transformations that map the input $X$ to queries $Q=XW_q$, keys $K=XW_k$ and values $V=XW_v$.
The outputs of all heads are then concatenated and projected again as follows:
\begin{equation}
\mha(X) = \concat\Big[O_1, \dots, O_{Nh}\Big]W^O
\label{eq:multihead}
\end{equation}
where $W^O \in \mathbb{R}^{d_v \times d_v}$ is a learned linear transformation.
$\mha(X)$ is then reshaped into a tensor of shape $(H,W,d_v)$ to match the original spatial dimensions.
We note that multi-head attention incurs a complexity of $O((HW)^2 d_k)$ and a memory cost of $O((H W)^2 N_h)$ as it requires to store attention maps for each head.

\subsubsection{Two-dimensional Positional Embeddings} \label{sec:pos-embedding}

Without explicit information about positions, self-attention is \emph{permutation equivariant}: 
$$\mha(\pi(X)) = \pi(\mha(X))$$ 
for any permutation $\pi$ of the pixel locations, making it ineffective for modeling highly structured data such as images.
Multiple positional encodings that augment activation maps with explicit spatial information have been proposed to alleviate related issues. 
In particular, the Image Transformer~\cite{parmar2018image} extends the sinusoidal waves first introduced in the original Transformer~\cite{vaswani2017attention} to 2 dimensional inputs and CoordConv~\cite{liu2018intriguing} concatenates positional channels to an activation map.

However these encodings did not help in our experiments on image classification and object detection (see Section~\ref{sec:ablation}). 
We hypothesize that this is because such positional encodings, while not permutation equivariant, do not satisfy \emph{translation equivariance}, which is a desirable property when dealing with images.
As a solution, we propose to extend the use of relative position encodings~\cite{shaw2018self} to two dimensions and present a memory efficient implementation based on the Music Transformer~\cite{huang2018improved}.

\paragraph{Relative positional embeddings:}
Introduced in \cite{shaw2018self} for the purpose of language modeling, relative self-attention augments self-attention with relative position embeddings and enables translation equivariance while preventing permutation equivariance. 
We implement two-dimensional relative self-attention by independently adding relative height information and relative width information.
The attention logit for how much pixel $i=(i_x, i_y)$ attends to pixel $j=(j_x,j_y)$ is computed as:

\begin{equation}
l_{i,j} = \frac{q_i^T}{\sqrt{d_k^h}} (k_j + r_{j_x - i_x}^W + r_{j_y - i_y}^H)
\label{eq:relative-att-unbatched}
\end{equation}
where $q_i$ is the query vector for pixel $i$ (the i-th row of $Q$), $k_j$ is the key vector for pixel $j$ (the j-th row of $K$) and $r_{j_x - i_x}^W$ and $r_{j_y - i_y}^H$ are learned embeddings for relative width $j_x - i_x$ and relative height $j_y - i_y$, respectively.
The output of head $h$ now becomes:
\begin{equation}
O_h = \softmax\left(\frac{QK^T + S_H^{rel} + S_W^{rel}}{\sqrt{d_k^h}} \right)V
\label{eq:relative-att}
\end{equation}
where $S_H^{rel}, S_W^{rel} \in \mathbb{R}^{HW \times HW}$ are matrices of relative position logits along height and width dimensions that satisfy $S_H^{rel}[i,j]  = q_i^T r_{j_y - i_y}^H$ and $S_W^{rel}[i,j] = q_i^T r_{j_x - i_x}^W$.

The relative attention algorithm in \cite{shaw2018self} explicitly stores all relative embeddings $r_{ij}$ in a tensor of shape $(HW, HW, d_k^h)$, thus incurring an additional memory cost of $O((HW)^2d_k^h)$.
This compares to $O((HW)^2N_h)$ for the position-unaware version self-attention that does not use position encodings. 
As we typically have $N_h < d_k^h$, such an implementation can prove extremely prohibitive and restrict the number of images that can fit in a minibatch. 
Instead, we extend the memory efficient relative masked attention algorithm presented in \cite{huang2018improved} to unmasked relative self-attention over 2 dimensional inputs. Our implementation has a memory cost of $O(HWd_k^h)$.
We leave the Tensorflow code of the algorithm in the Appendix.

The relative positional embeeddings $r^H$ and $r^W$ are learned and shared across heads but not layers. For each layer, we add $(2(H+W)-2)d_k^h$ parameters to model relative distances along height and width.

\subsection{Attention Augmented Convolution} \label{sec:augmented-conv}

Multiple previously proposed attention mechanisms over images \cite{hu2017squeeze,hu2018gather,park2018bam,woo2018cbam} suggest that the convolution operator is limited by its locality and lack of understanding of global contexts. These methods capture long-range dependencies by recalibrating convolutional feature maps. In particular, Squeeze-and-Excitation (SE) \cite{hu2017squeeze} and Gather-Excite (GE) \cite{hu2018gather} perform channelwise reweighing while BAM \cite{park2018bam} and CBAM \cite{woo2018cbam} reweigh both channels and spatial positions \emph{independently}.
In contrast to these approaches, we 1) use an attention mechanism that can attend \emph{jointly} to spatial and feature subspaces (each head corresponding to a feature subspace) and 2) introduce additional feature maps rather than refining them. Figure \ref{fig:attention-diagram} summarizes our proposed augmented convolution.

\paragraph{Concatenating convolutional and attentional feature maps:}

Formally, consider an original convolution operator with kernel size $k$, $F_{in}$ input filters and $F_{out}$ output filters. The corresponding attention augmented convolution can be written as 
$$\aaconv(X) = \concat\Big[\conv(X),\mha(X)\Big].$$
We denote $\upsilon = \frac{d_v}{F_{out}}$ the ratio of attentional channels to number of original output filters and $\kappa = \frac{d_k}{F_{out}}$ the ratio of key depth to number of original output filters.
Similarly to the convolution, the proposed attention augmented convolution 1) is equivariant to translation and 2) can readily operate on inputs of different spatial dimensions.
We include Tensorflow code for the proposed attention augmented convolution in the Appendix \ref{sec:code}.

\paragraph{Effect on number of parameters:}
Multihead attention introduces a 1x1 convolution with $F_{in}$ input filters and $(2d_k + d_v) = F_{out} (2 \kappa + \upsilon)$ output filters to compute queries, keys and values and an additional 1x1 convolution with $d_v = F_{out} \upsilon$ input and output filters to mix the contribution of different heads.
Considering the decrease in filters in the convolutional part, this leads to the following change in parameters:
\begin{equation}
\Delta_{params} \sim F_{in} F_{out} (2 \kappa + (1 - k^2) \upsilon + \frac{F_{out}}{F_{in}} \upsilon^2),
\end{equation}\label{eq:params}
where we ignore the parameters introduced by relative position embeddings for simplicity as these are negligible.
In practice, this causes a slight decrease in parameters when replacing 3x3 convolutions and a slight increase in parameters when replacing 1x1 convolutions. Interestingly, we find in experiments that attention augmented networks still significantly outperform their fully convolutional counterparts while using less parameters.

\paragraph{Attention Augmented Convolutional Architectures:}

In all our experiments, the augmented convolution is followed by a batch normalization~\cite{BatchNorm} layer which can learn to scale the contribution of the convolution feature maps and the attention feature maps. 
We apply our augmented convolution once per residual block similarly to other visual attention mechanisms~\cite{hu2017squeeze,hu2018gather,park2018bam,woo2018cbam} and along the entire architecture as memory permits (see Section \ref{sec:experiments} for more details).

Since the memory cost $O((N_h(HW)^2)$ can be prohibitive for large spatial dimensions, we augment convolutions with attention starting from the last layer (with smallest spatial dimension) until we hit memory constraints. To reduce the memory footprint of augmented networks, we typically resort to a smaller batch size and sometimes additionally downsample the inputs to self-attention in the layers with the largest spatial dimensions where it is applied. Downsampling is performed by applying 3x3 average pooling with stride 2 while the following upsampling (required for the concatenation) is obtained via bilinear interpolation.

%% file: results.tex
\section{Experiments} \label{sec:experiments}

In the subsequent experiments, we test Attention Augmentation on standard computer vision architectures such as ResNets~\cite{identity-mappings, xie2016aggregated, he2015deep}, and MnasNet~\cite{tan2018mnasnet} on the CIFAR-100 \cite{krizhevsky2009learning}, ImageNet \cite{deng2009imagenet} and COCO \cite{lin2016feature} datasets. 
Our experiments show that Attention Augmentation leads to systematic improvements on both image classification and object detection tasks across a broad array of architectures and computational demands. We validate the utility of the proposed two-dimensional relative attention mechanism in ablation experiments.
In all experiments, we substitute convolutional feature maps with self-attention feature maps as it makes for an easier comparison against the baseline models. Unless specified otherwise, all results correspond to our two-dimensional relative self-attention mechanism.
Experimental details can be found in the Appendix.

\subsection{CIFAR-100 image classification}

We first investigate how Attention Augmentation performs on CIFAR-100 \cite{krizhevsky2009learning}, a standard benchmark for low-resolution imagery, using a Wide ResNet architecture~\cite{wide}.
The Wide-ResNet-28-10 architecture is comprised of 3 stages of 4 residual blocks each using two $3\times3$ convolutions.
We augment the Wide-ResNet-28-10 by augmenting the first convolution of all residual blocks with relative attention using $N_h$=8 heads and $\kappa$=$2\upsilon$=0.2 and a minimum of 20 dimensions per head for the keys. 
We compare Attention Augmentation (AA) against other forms of attention including Squeeze-and-Excitation (SE) \cite{hu2017squeeze} and the parameter-free formulation of Gather-Excite (GE) \cite{hu2018gather}.
Table \ref{tab:cifar100_experiments} shows that Attention Augmentation improves performance both over the baseline network and Squeeze-and-Excitation at a similar parameter and complexity cost.

\begin{table}[h!]
\begin{center}
{\small
\begin{tabular}{|l|c|c|c|c|}
\hline
Architecture & Params & GFlops & top-1 & top-5 \\
\hline
Wide-ResNet \cite{wide} & 36.3M & 10.4 & 80.3 & 95.0 \\
GE-Wide-ResNet \cite{hu2018gather} & 36.3M & 10.4 & 79.8 & 95.0 \\
SE-Wide-ResNet \cite{hu2017squeeze} & 36.5M & 10.4 & 81.0 & 95.3 \\
AA-Wide-ResNet (ours) & 36.2M & 10.9 & 81.6 & 95.2 \\
\hline
\end{tabular}
} 
\end{center}
\caption{Image classification on the CIFAR-100 dataset \cite{krizhevsky2009learning} using the Wide-ResNet 28-10 architecture \cite{wide}.}
\label{tab:cifar100_experiments}
\end{table}

\subsection{ImageNet image classification with ResNet} \label{sec:resnet_experiments}

We next examine how Attention Augmentation performs on ImageNet \cite{deng2009imagenet, kornblith2019better}, a standard large-scale dataset for high resolution imagery, across an array of architectures.
We start with the ResNet architecture~\cite{identity-mappings, xie2016aggregated, he2015deep} because of its widespread use and its ability to easily scale across several computational budgets.
The building block in ResNet-34 comprises two 3x3 convolutions with the same number of output filters. 
ResNet-50 and its larger counterparts use a bottleneck block comprising of 1x1, 3x3, 1x1 convolutions where the last pointwise convolution expands the number of filters and the first one contracts the number of filters. 
We modify all ResNets by augmenting the 3x3 convolutions as this decreases number of parameters.\footnote{We found that augmenting the pointwise expansions works just as well but does not save parameters or computations.}
We apply Attention Augmentation in each residual block of the last 3 stages of the architecture -- when the spatial dimensions of the activation maps are 28x28, 14x14 and 7x7 -- and downsample only during the first stage.
All attention augmented networks use  $\kappa$=$2\upsilon$=0.2, except for ResNet-34 which uses $\kappa$=$\upsilon$=0.25. 
The number of attention heads is fixed to $N_h$=8.

\begin{table}[h!]
\begin{center}
{\footnotesize
\begin{tabular}{|l|c|c|c|c|}
\hline
Architecture & Params (M) & $\Delta_{Infer}$ & $\Delta_{Train}$ & top-1 \\
\hline
ResNet-50 & 25.6 & - & - & 76.4 \\
SE \cite{hu2017squeeze} & 28.1 & +12\% & +92\% & 77.5 (77.0) \\
BAM \cite{park2018bam} & 25.9 & +19\% & +43\% & 77.3 \\
CBAM \cite{woo2018cbam} & 28.1 & +56\% & +132\% & 77.4 (77.4) \\
GALA \cite{linsley2018gala} & 29.4 & +86\% & +133\% & 77.5 (77.3) \\
\textbf{AA ($\upsilon=0.25$)} & \textbf{24.3} & \textbf{+29\%} & \textbf{+25\%} & \textbf{77.7} \\
\hline
\end{tabular}
} 
\end{center}
\caption{Image classification performance of different attention mechanisms on the ImageNet dataset.
$\Delta$ refers to the increase in latency times compared to the ResNet50 on a single Tesla V100 GPU with Tensorflow using a batch size of 128.
For fair comparison, we also include top-1 results (in parentheses) when scaling networks in width to match $\sim25.6$M parameters as the ResNet50 baseline.}
\label{tab:attention_compare}
\end{table}

\begin{table}[h!]
\begin{center}
{\small
\begin{tabular}{|l|c|c|c|c|}
\hline
Architecture & GFlops & Params & top-1 & top-5 \\
\hline
ResNet-34 \cite{identity-mappings} & 7.4 & 21.8M & 73.6 & 91.5 \\
SE-ResNet-34 \cite{hu2017squeeze} & 7.4 & 22.0M & 74.3 & 91.8 \\
AA-ResNet-34 (ours) & 7.1 & 20.7M & 74.7 & 92.0 \\
\hline
ResNet-50 \cite{identity-mappings} & 8.2 & 25.6M & 76.4 & 93.1 \\
SE-ResNet-50 \cite{hu2017squeeze} & 8.2 & 28.1M & 77.5 & 93.7 \\
AA-ResNet-50 (ours) & 8.3 & 25.8M & 77.7 & 93.8 \\
\hline
ResNet-101 \cite{identity-mappings} & 15.6 & 44.5M & 77.9 & 94.0 \\
SE-ResNet-101 \cite{hu2017squeeze} & 15.6 & 49.3M & 78.4 & 94.2 \\
AA-ResNet-101 (ours) & 16.1 & 45.4M & 78.7 & 94.4 \\
\hline
ResNet-152 \cite{identity-mappings} & 23.0 & 60.2M & 78.4 & 94.2 \\
SE-ResNet-152 \cite{hu2017squeeze} & 23.1 & 66.8M & 78.9 & 94.5 \\
AA-ResNet-152 (ours) & 23.8 & 61.6M & 79.1 & 94.6 \\
\hline
\end{tabular}
} 
\end{center}
\caption{Image classification on the ImageNet dataset \cite{deng2009imagenet} across a range of ResNet architectures: ResNet-34, ResNet-50, Resnet-101, and ResNet-152 \cite{identity-mappings, xie2016aggregated, he2015deep}.
}
\label{tab:resnet_experiments}
\end{table}

Table \ref{tab:attention_compare} benchmarks Attention Augmentation against \emph{channel and spatial attention} mechanisms BAM~\cite{park2018bam}, CBAM~\cite{woo2018cbam} and GALA~\cite{linsley2018gala} with channel reduction ratio $\sigma=16$ on the ResNet50 architecture.
Despite the lack of specialized kernels (See Appendix \ref{sec:code}), Attention Augmentation offers a competitive accuracy/computational trade-off compared to previously proposed attention mechanisms.
Table~\ref{tab:resnet_experiments} compares the non-augmented networks and Squeeze-and-Excitation (SE)~\cite{hu2017squeeze} across different network scales.
In all experiments, Attention Augmentation significantly increases performance over the non-augmented baseline and notably outperforms Squeeze-and-Excitation (SE)~\cite{hu2017squeeze} while being more parameter efficient (Figure~\ref{fig:marketing}). 
Remarkably, our AA-ResNet-50 performs comparably to the baseline ResNet-101 and our AA-ResNet-101 outperforms the baseline ResNet-152.
These results suggest that attention augmentation is preferable to simply making networks deeper.
We include and discuss attention maps visualizations from different pixel positions in the appendix.

\subsection{ImageNet classification with MnasNet}

\begin{table}[t!]
\begin{center}
{\small
\begin{tabular}{|l|c|c|c|c|}
\hline
Architecture & GFlops & Params & top-1 & top-5  \\
\hline
MnasNet-0.75 & 0.45 & 2.91M & 73.3 & 91.3  \\
AA-MnasNet-0.75 & 0.51 & 3.02M & 73.9 & 91.6  \\
\hline
MnasNet-1.0 & 0.63 & 3.89M & 75.2 & 92.4 \\
AA-MnasNet-1.0 & 0.70 & 4.06M & 75.7 & 92.6 \\
\hline
MnasNet-1.25 & 1.01 & 5.26M & 76.7 & 93.2  \\
AA-MnasNet-1.25 & 1.11 & 5.53M & 77.2 & 93.6  \\
\hline
MnasNet-1.4 & 1.17 & 6.10M & 77.2 & 93.5  \\
AA-MnasNet-1.4 & 1.29 & 6.44M & 77.7 & 93.8  \\
\hline
\end{tabular}
} 
\end{center}
\caption{Baseline and attention augmented MnasNet \cite{tan2018mnasnet} accuracies with width multipliers 0.75, 1.0, 1.25 and 1.4.}
\label{tab:MnasNet_experiments}
\end{table}

\begin{table*}[t!]
\begin{center}
\begin{tabular}{|l|c|c|c|c|c|}
\hline
Backbone architecture & GFlops & Params & mAP$_{\sf COCO}$ & mAP$_{50}$ & mAP$_{75}$ \\
\hline
ResNet-50 \cite{lin2017focal} & 182 & 33.4M & 36.8 & 54.5 & 39.5 \\
SE-ResNet-50 \cite{hu2017squeeze} & 183 & 35.9M & 36.5 & 54.0 & 39.1  \\
AA-ResNet-50 (ours)  & 182 & 33.1M & 38.2 & 56.5 & 40.7 \\
\hline
ResNet-101 \cite{lin2017focal} & 243 & 52.4M & 38.5 & 56.4 & 41.2 \\
SE-ResNet-101 \cite{hu2017squeeze} & 243 & 57.2M & 37.4 & 55.0 & 39.9  \\
AA-ResNet-101 (ours)  & 245 & 51.7M & 39.2 & 57.8 & 41.9 \\
\hline
\end{tabular}
\end{center}
\caption{Object detection on the COCO dataset \cite{lin2014microsoft} using the RetinaNet architecture \cite{lin2017focal} with different backbone architectures. We report mean Average Precision at three different IoU values.}
\label{tab:retinanet_experiments}
\end{table*}

In this section, we inspect the use of Attention Augmentation in a resource constrained setting by conducting ImageNet experiments with the MnasNet architecture~\cite{tan2018mnasnet}, which is an extremely parameter-efficient architecture. 
In particular, the MnasNet was found by neural architecture search~\cite{zoph2017neural}, using only the highly optimized mobile inverted bottleneck block~\cite{sandler2018mobilenetv2} and the Squeeze-and-Excitation operation~\cite{hu2017squeeze} as the primitives in its search space.
We apply Attention Augmentation to the mobile inverted bottleneck by replacing convolutional channels in the expansion pointwise convolution using  $\kappa$=$2\upsilon$=0.1 and $N_h$=4 heads.
Our augmented MnasNets use augmented inverted bottlenecks in the last 13 blocks out of 18 in the MnasNet architecture, starting when the spatial dimension is 28x28. We downsample only in the first stage where Attention Augmentation is applied. We leave the final pointwise convolution, also referred to as the ``head", unchanged.

\begin{figure}[t!]
    \centering
    \includegraphics[width=1.0\columnwidth]{{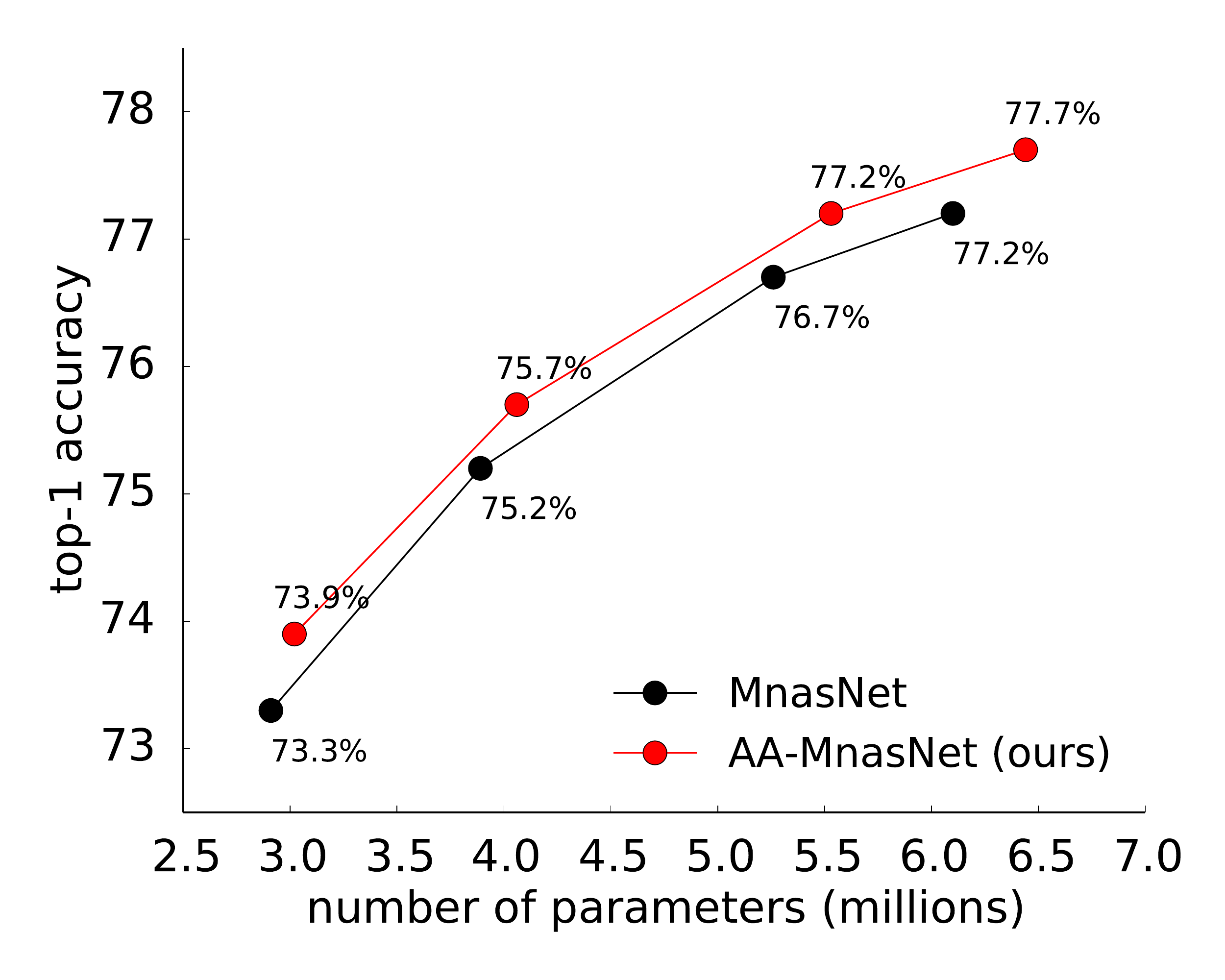}}
    \caption{ImageNet top-1 accuracy as a function of number of parameters for MnasNet (black) and Attention-Augmented-MnasNet (red) with width multipliers $0.75$, $1.0$, $1.25$ and $1.4$.}
    \label{fig:MnasNet_params_curve}
\end{figure}

In Table \ref{tab:MnasNet_experiments}, we report ImageNet accuracies for the baseline MnasNet and its attention augmented variants at different width multipliers. 
Our experiments show that Attention Augmentation yields accuracy improvements across all width multipliers.
Augmenting MnasNets with relative self-attention incurs a slight parameter increase, however we verify in Figure \ref{fig:MnasNet_params_curve} that the accuracy improvements are not just explained by the parameter increase.
Additionally, we note that the MnasNet architecture employs Squeeze-and-Excitation at multiple locations that were optimally selected via architecture search, further suggesting the benefits of our method.

\subsection{Object Detection with COCO dataset}

We next investigate the use of Attention Augmentation on the task of object detection on the COCO dataset~\cite{lin2014microsoft}.
We employ the RetinaNet architecture with a ResNet-50 and ResNet-101 backbone as done in \cite{lin2017focal}, using the opensourced RetinaNet codebase.\footnote{\url{https://github.com/tensorflow/tpu/tree/master/models/official/retinanet}}
We apply Attention Augmentation uniquely on the ResNet backbone, modifying them similarly as in our ImageNet classification experiments.

Our relative self-attention mechanism improves the performance of the RetinaNet on both ResNet-50 and ResNet-101 as shown in Table~\ref{tab:retinanet_experiments}. 
Most notably, Attention Augmentation yields a 1.4\% mAP improvement over a strong RetinaNet baseline from~\cite{lin2017focal}. 
In contrast to the success of Squeeze-and-Excitation in image classification with ImageNet, our experiments show that adding Squeeze-and-Excitation operators in the backbone network of the RetinaNet significantly hurts performance, in spite of grid searching over the squeeze ratio $\sigma \in \{4, 8, 16\}$.
We hypothesize that localization requires precise spatial information which SE discards during the spatial pooling operation, thereby negatively affecting performance. 
Self-attention on the other hand maintains spatial information and is likely to be able to identify object boundaries successfully. Visualizations of attention maps (See Figures~\ref{fig:input_image} and~\ref{fig:visualization} in the Appendix) reveal that some heads are indeed delineating objects from their background which might be important for localization.

\subsection{Ablation Study} \label{sec:ablation}

\paragraph{Fully-attentional vision models:}

In this section, we investigate the performance of Attention Augmentation as a function of the fraction of attentional channels. 
As we increase this fraction to 100\%, we begin to replace a ConvNet with a fully attentional model, only leaving pointwise convolutions and the stem unchanged.
Table \ref{tab:ablation_experiments} presents the performance of Attention Augmentation on the ResNet-50 architecture for varying ratios $\kappa$=$\upsilon$ $\in \{0.25, 0.5, 0.75, 1.0\}$.
Performance slightly degrades as the ratio of attentional channels increases, which we hypothesize is partly explained by the average pooling operation for downsampling at the first stage where Attention Augmentation is applied.
Attention Augmentation proves however quite robust to the fraction of attentional channels.
For instance, AA-ResNet-50 with $\kappa$=$\upsilon$=0.75 outperforms its ResNet-50 counterpart, while being more parameter and flops efficient, indicating that mostly employing attentional channels is readily competitive.

Perhaps surprisingly, these experiments also reveal that our proposed self-attention mechanism is a powerful stand-alone computational primitive for image classification and that fully attentional models are viable for discriminative visual tasks.
In particular, AA-ResNet-50 with $\kappa$=$\upsilon$=1, which uses exclusively attentional channels, is only 2.5\% worse in accuracy than its fully convolutional counterpart, in spite of downsampling with average pooling and having 25\% less parameters.
Notably, this fully attentional architecture\footnote{We consider pointwise convolutions as dense layers. This architecture employs 4 non-pointwise convolutions in the stem and the first stage of the architecture, but we believe such operations can be replaced by attention too.} also outperforms ResNet-34 while being more parameter and flops efficient (see Table \ref{tab:ablation_experiments}).

\begin{table}[h!]
\begin{center}
\begin{tabular}{|l|c|c|c|c|}
\hline
Architecture & GFlops & Params & top-1 & top-5 \\
\hline
ResNet-34 \cite{identity-mappings} & 7.4 & 21.8M & 73.6 & 91.5 \\
ResNet-50 \cite{identity-mappings} & 8.2 & 25.6M & 76.4 & 93.1 \\
\hline
$\kappa=\upsilon=0.25$ & 7.9 & 24.3M & 77.7 & 93.8 \\
$\kappa=\upsilon=0.5$ & 7.3 & 22.3M & 77.3 & 93.6 \\
$\kappa=\upsilon=0.75$ & 6.8 & 20.7M & 76.7 & 93.2 \\
$\kappa=\upsilon=1.0$ & 6.3 & 19.4M & 73.9 & 91.5 \\
\hline
\end{tabular}
\end{center}
\caption{Attention Augmented ResNet-50 with varying ratios of attentional channels.}
\label{tab:ablation_experiments}
\end{table}

\begin{figure}[t!]
    \centering
    \includegraphics[width=1.0\columnwidth]{{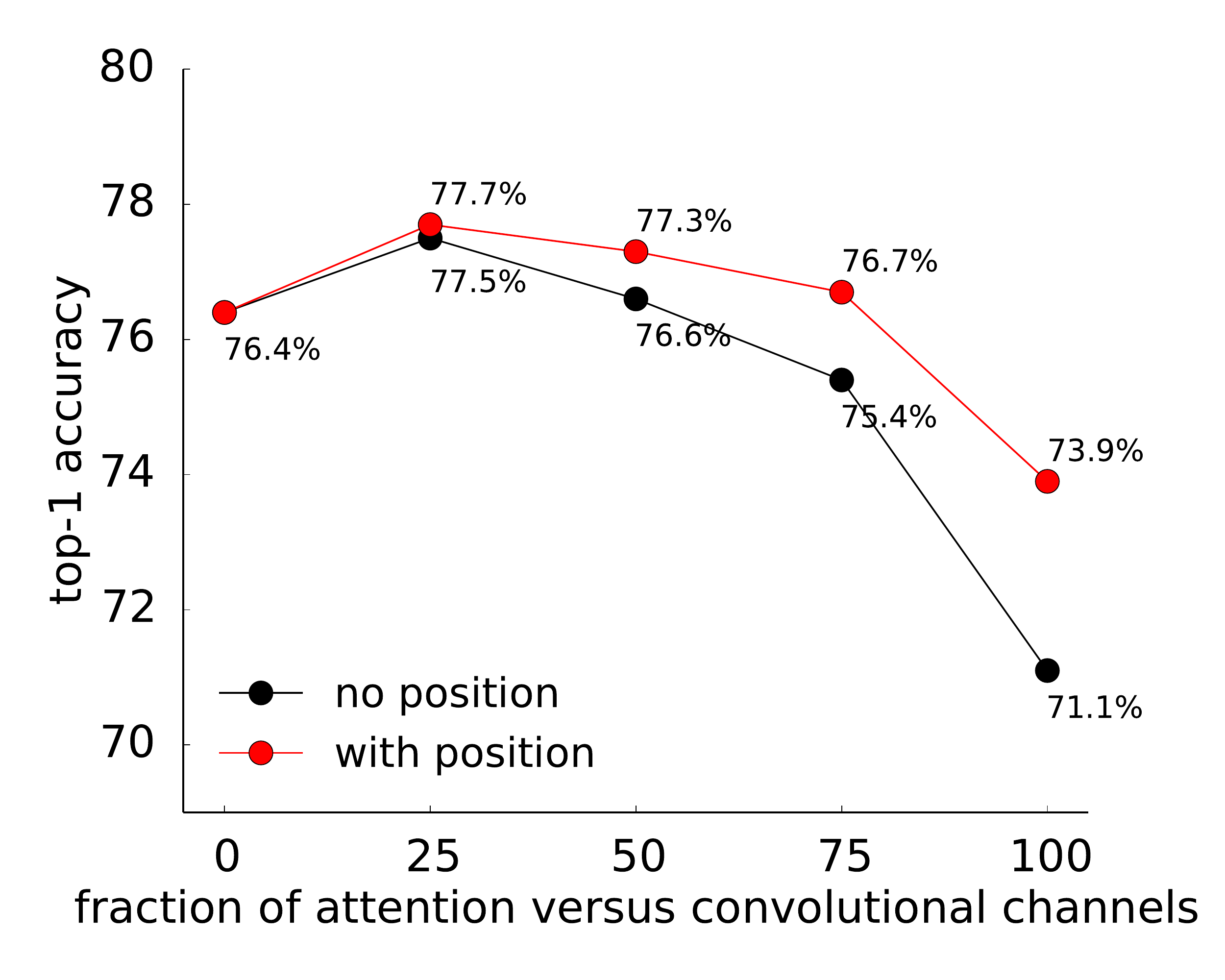}}    
    \caption{Effect of relative position embeddings as the ratio of attentional channels increases on our Attention-Augmented ResNet50.}
    \label{fig:ablation}
\end{figure}

\paragraph{Importance of position encodings:}

In Figure \ref{fig:ablation}, we show the effect of our proposed two-dimensional relative position encodings as a function of the fraction of attentional channels. 
As expected, experiments demonstrate that our relative position encodings become increasingly more important as the architecture employs more attentional channels. 
In particular, the fully self-attentional ResNet-50 gains 2.8\% top-1 ImageNet accuracy when using relative position encodings, which indicates the necessity of maintaining position information for fully self-attentional vision models.

We additionally compare our proposed two-dimensional relative position encodings to other position encoding schemes.
We apply Attention Augmentation using the same hyperparameters as \ref{sec:resnet_experiments} with the following different position encoding schemes: 1) The position-unaware version of self-attention (referred to as \textit{None}), 2) a two-dimensional implementation of the sinusoidal positional waves  (referred to as \textit{2d Sine}) as used in~\cite{parmar2018image}, 3) CoordConv~\cite{liu2018intriguing} for which we concatenate (x,y,r) coordinate channels to the inputs of the attention function, and 4) our proposed two-dimensional relative position encodings (referred to as \textit{Relative}).

\begin{table}[t!]
\begin{center}
\begin{tabular}{|c|c|c|c|c|}
\hline
Architecture & Position Encodings & top-1 & top-5 \\
\hline
AA-ResNet-34 & None & 74.4 & 91.9 \\
AA-ResNet-34 & 2d Sine & 74.4 & 92.0 \\
AA-ResNet-34 & CoordConv & 74.4 & 92.0 \\
AA-ResNet-34 & Relative (ours) & 74.7 & 92.0 \\
\hline
AA-ResNet-50 & None & 77.5 & 93.7 \\
AA-ResNet-50 & 2d Sine & 77.5 & 93.7 \\
AA-ResNet-50 & CoordConv & 77.5 & 93.8 \\
AA-ResNet-50 & Relative (ours) & 77.7 & 93.8 \\
\hline
\end{tabular}
\end{center}
\caption{Effects of different position encodings in Attention Augmentation on ImageNet classification.}
\label{table:resnet-ablation}
\end{table}

\begin{table}[t!]
\begin{center}
\begin{tabular}{|l|c|c|c|c|c|}
\hline
Position Encodings & mAP$_{\sf COCO}$ & mAP$_{50}$ & mAP$_{75}$ \\
\hline
None & 37.7 & 56.0 & 40.2 \\
CoordConv \cite{liu2018intriguing} & 37.4 & 55.5 & 40.1 \\
Relative (ours) & 38.2 & 56.5 & 40.7 \\
\hline
\end{tabular}
\end{center}
\caption{Effects of different position encodings in Attention Augmentation on the COCO object detection task using a RetinaNet AA-ResNet-50 backbone.}
\label{table:detection-ablation}
\end{table}

In Table \ref{table:resnet-ablation} and \ref{table:detection-ablation}, we present the results on ImageNet classification and the COCO object detection task respectively.
On both tasks, Attention Augmentation without position encodings already yields improvements over the fully convolutional non-augmented variants. 
Our experiments also reveal that the sinusoidal encodings and the coordinate convolution do not provide improvements over the position-unaware version of Attention Augmentation. 
We obtain additional improvements when using our two-dimensional relative attention, demonstrating the utility of preserving translation equivariance while preventing permutation equivariance.

%% file: discussion.tex
\section{Discussion and future work}

In this work, we consider the use of self-attention for vision models as an alternative to convolutions.
We introduce a novel two-dimensional relative self-attention mechanism for images that enables training of competitive fully self-attentional vision models on image classification for the first time.
We propose to augment convolutional operators with this self-attention mechanism and validate the superiority of this approach over other attention schemes.
Extensive experiments show that Attention Augmentation leads to systematic improvements on both image classification and object detection tasks across a wide range of architectures and computational settings.

Several open questions from this work remain. 
In future work, we will focus on the fully attentional regime and explore how different attention mechanisms trade off computational efficiency versus representational power.
For instance, identifying a {\em local} attention mechanism may result in an efficient and scalable computational mechanism that could prevent the need for downsampling with average pooling \cite{ramachandran2019sasa}.
Additionally, it is plausible that architectural design choices that are well suited when exclusively relying on convolutions are suboptimal when using self-attention mechanisms. 
As such, it would be interesting to see if using Attention Augmentation as a primitive in automated architecture search procedures proves useful to find even better models than those previously found in image classification~\cite{zoph2018learning}, object detection~\cite{ghiasi2019NASFpn}, image segmentation~\cite{chen2018searching} and other domains~\cite{bello2017nos,alber2018backprop,ramachadran2017searching,dogus2018autoaugment}.
Finally, one can ask to which degree fully attentional models can replace convolutional networks for visual tasks.

%% file: appendix.tex
\clearpage

\appendix

\section{Appendix}

\subsection{Experimental details}

\paragraph{Tuning}
Unless specified otherwise, we use the default hyperparameters found in reference baseline codebases without tuning.
$\kappa$ was searched in \{0.1, 0.2, 0.5\}, $\upsilon$ in \{0.0, 0.1, 0.25, 0.5, 0.75, 1.0\} and the number of heads was chosen based on memory constraints (starting from 8 and decreasing when necessary).
We report the final accuracy for each run without performing early stopping.

\vspace{-5pt}
\paragraph{CIFAR-100}
Given the low resolution of CIFAR-100 images, we do not downsample feature maps before the attention operation and instead resort to a smaller batch size.
We train all networks for 500 epochs using synchronous SGD with momentum 0.9 distributed across 8 TESLA V100 GPUs. The learning rate is linearly scaled from 0 to $0.2B/256$, where $B$ is the total batch size, for the first $5\%$ training epochs and then annealed with cosine decay~\cite{cosine}.
We use standard CIFAR preprocessing: mean normalizing, random flipping and cropping~\cite{zoph2018learning,gastaldi2017shake,yamada2018shakedrop}.
Non-augmented architectures are trained with a batch size of 1024 and a weight decay of 2e-4. Augmented architectures are trained with batch size of 256 and a weight decay of 5e-4.

\vspace{-5pt}
\paragraph{ImageNet classification with ResNet}
We train all ResNet architectures for 100 epochs using synchronous SGD with momentum 0.9 across 8 TESLA V100 GPUs and weight decay of 1e-4. 
We use the largest batch size per worker $B \in \{32, 64, 128, 256\}$ that fits in a minibatch.
The initial learning rate is scaled linearly according to the total batch size using a base learning rate of 0.128 for total batch size of 256. 
During training, we linearly scale the learning rate from 0 to this value for the first 5\% of training epochs and divide it by 10 at epochs 30, 60, 80 and 90. 
We use standard Inception data augmentation as described in~\cite{szegedy2016rethinking}.

\vspace{-5pt}
\paragraph{ImageNet classification with MnasNet}
We follow the training setup described in~\cite{tan2018mnasnet} and train all networks for 350 epochs with the RMSProp optimizer using exponential learning rate decay. 
When training our augmented MnasNets, we divide the learning rate by 2 and adjusted the learning rate decay so that the final learning rate stays the same.

\vspace{-5pt}
\paragraph{Object Detection with COCO dataset}
We follow the setup described in \cite{lin2017focal,ghiasi2018dropblock} and train the RetinaNet from scratch for 150 epochs without using ImageNet pretraining for the ResNet backbone. 

\subsection{Computational \& Memory costs}
Table \ref{tab:memory_cost} provides the breakdown of self-attention related computational costs per image.
Storing attention maps in each layer induces a memory cost of $N_h (HW)^2$ {\tt bfloat16}.
At inference, the memory cost for storing attention maps is only 1.2\% of the memory required to store model parameters (49MB).

\begin{table}[h!]
\begin{center}
{\small
\begin{tabular}{|l|c|c|c|}
\hline
Layer & Memory & Params & FLOPS \\
\hline
\{Stage 2 - H=W=14\} * 4 & 600KB & 43k & 22M \\
\{Stage 3 - H=W=14\} * 6 & 600KB & 90k & 40M \\
\{Stage 4 - H=W=7\} * 3 & 37.5KB & 190k & 19M \\
\hline
\emph{Training} & 6MB (total) & 1.3M & 390M \\
\hline
\emph{Inference} & 600KB (max) & 1.3M & 390M \\
\hline
\end{tabular}
} 
\end{center}
\caption{Computational costs associated with self-attention in the forward pass of the ResNet50.
During inference, we only consider the largest memory cost since activations are not stored.
}
\label{tab:memory_cost}
\end{table}

\begin{figure}[h!]
    \centering
    \includegraphics[width=0.8\columnwidth]{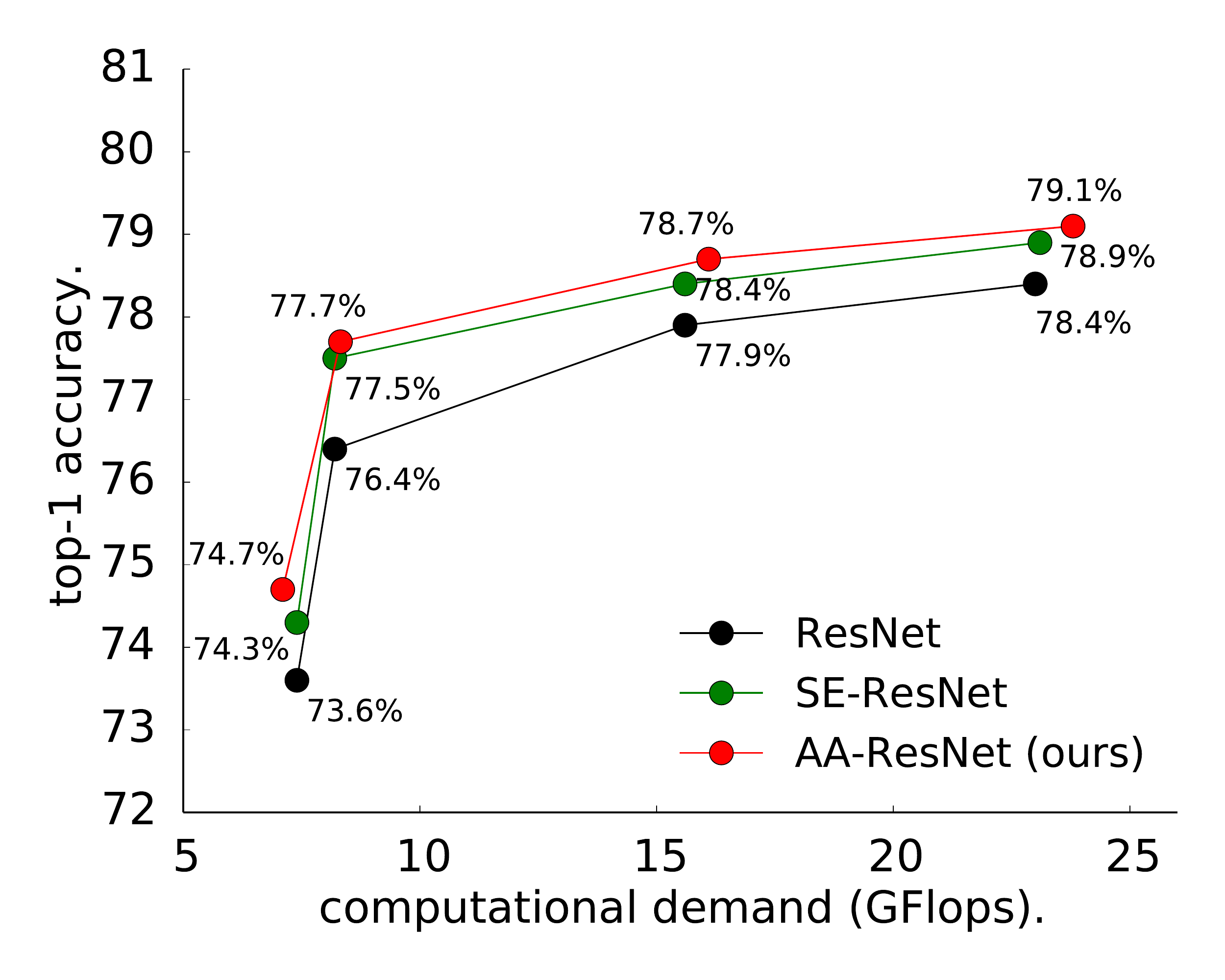}
    \caption{ImageNet top-1 accuracy as a function of computational demand for variety of ResNet architectures \cite{identity-mappings}. From left to right: ResNet-34, ResNet-50, ResNet-101 and ResNet-152.}
    \label{fig:resnet_curve_flops}
    \vspace{-5mm}
\end{figure}

\begin{figure}[h!]
    \centering
    \includegraphics[width=0.8\columnwidth]{{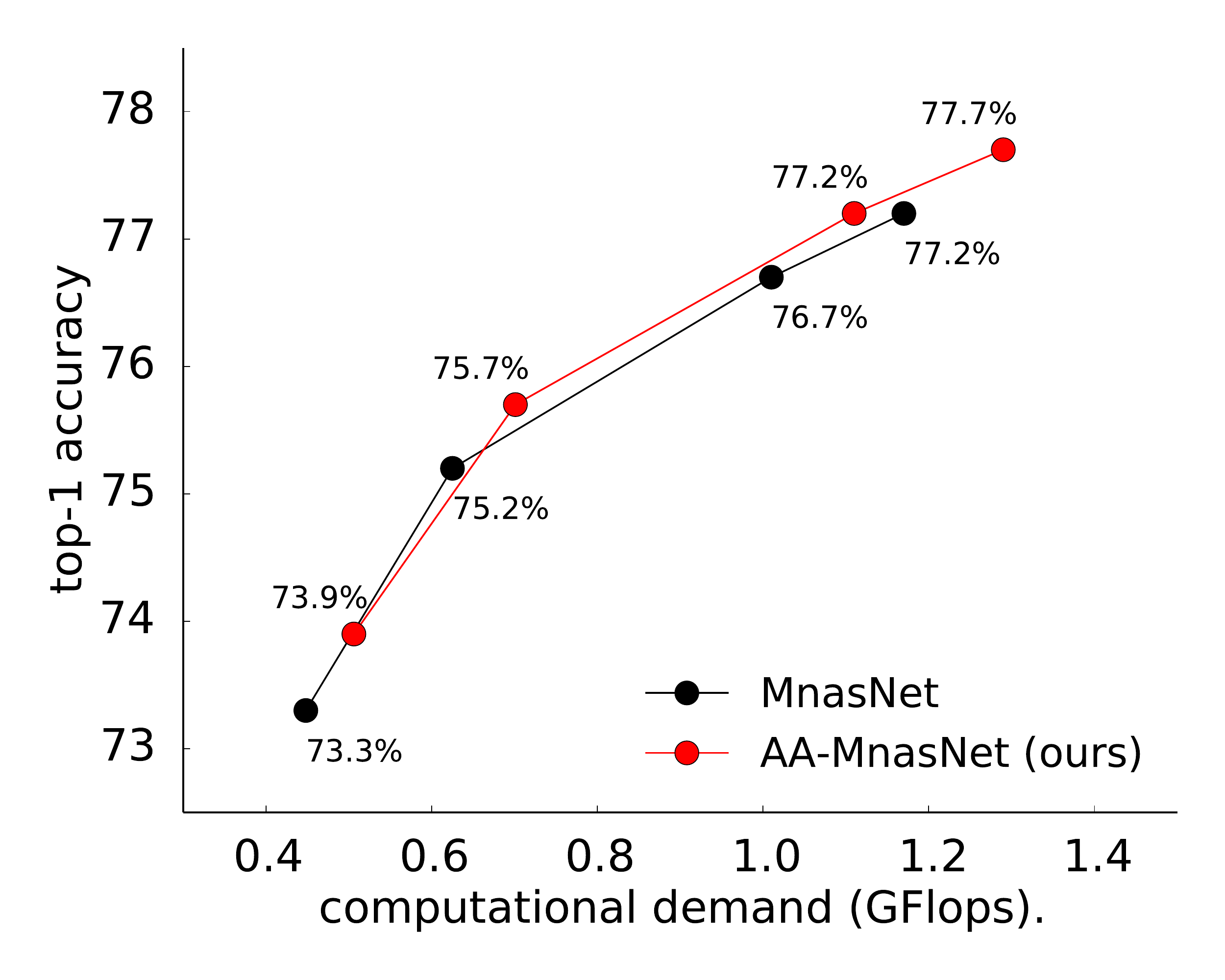}}
    \caption{ImageNet top-1 accuracy as a function of computational demand for MnasNet (black) and Attention-Augmented-MnasNet (red) with width multipliers $0.75$, $1.0$, $1.25$ and $1.4$.}
    \label{fig:mnasnet_curve_flops}
\end{figure}

Figures~\ref{fig:resnet_curve_flops} and~\ref{fig:mnasnet_curve_flops} show the accuracies of our attention augmented networks across FLOPS counts, which correlate with running times across hardware platforms.

\subsection{2D Relative Self-Attention implementation\label{sec:code}}

While our method is simple and only requires matrix multiplication, addition and the softmax operation (Equations \ref{eq:relative-att-unbatched} and \ref{eq:relative-att}), our implementation relies on non-trivial operations (e.g. tiling, transposing and reshaping) because no low-level kernels currently exist for hardware platforms.
Future work may develop specialized kernels as previously done for convolutions.
Therefore, we believe that current latency times (Table \ref{tab:attention_compare}) reflect the lack of dedicated engineering as opposed to inefficiency in the proposed method.

\lstset{
  backgroundcolor=\color{white},
  basicstyle=\fontsize{7.5pt}{8.5pt}\fontfamily{lmtt}\selectfont,
  columns=fullflexible,
  breaklines=true,
  captionpos=b,
  commentstyle=\fontsize{8pt}{9pt}\color{codegray},
  keywordstyle=\fontsize{8pt}{9pt}\color{codegreen},
  stringstyle=\fontsize{8pt}{9pt}\color{codeblue},
  frame=tb,
  otherkeywords = {self},
}
\begin{figure}[h!]
\tiny
\begin{lstlisting}[language=python]
def shape_list(x):
  """Return list of dims, statically where possible."""
  static = x.get_shape().as_list()
  shape = tf.shape(x)
  ret = []
  for i, static_dim in enumerate(static):
    dim = static_dim or shape[i]
    ret.append(dim)
  return ret

def split_heads_2d(inputs, Nh):
  """Split channels into multiple heads."""
  B, H, W, d = shape_list(inputs)
  ret_shape = [B, H, W, Nh, d // Nh]
  split = tf.reshape(inputs, ret_shape)
  return tf.transpose(split, [0, 3, 1, 2, 4])

def combine_heads_2d(inputs):
  """Combine heads (inverse of split_heads_2d)."""
  transposed = tf.transpose(inputs, [0, 2, 3, 1, 4])
  Nh, channels = shape_list(transposed)[-2:]
  ret_shape = shape_list(transposed)[:-2] + [Nh * channels]
  return tf.reshape(transposed, ret_shape)

def rel_to_abs(x):
  """Converts tensor from relative to aboslute indexing."""
  # [B, Nh, L, 2L-1]
  B, Nh, L, _ = shape_list(x)
  # Pad to shift from relative to absolute indexing.
  col_pad = tf.zeros((B, Nh, L, 1))
  x = tf.concat([x, col_pad], axis=3)
  flat_x = tf.reshape(x, [B, Nh, L * 2 * L])
  flat_pad = tf.zeros((B, Nh, L-1))
  flat_x_padded = tf.concat([flat_x, flat_pad], axis=2)
  # Reshape and slice out the padded elements.
  final_x = tf.reshape(flat_x_padded, [B, Nh, L+1, 2*L-1])
  final_x = final_x[:, :, :L, L-1:]
  return final_x

def relative_logits_1d(q, rel_k, H, W, Nh, transpose_mask):
  """Compute relative logits along one dimenion."""
  rel_logits = tf.einsum('bhxyd,md->bhxym', q, rel_k)
  # Collapse height and heads
  rel_logits = tf.reshape(
      rel_logits, [-1, Nh * H, W, 2 * W-1])
  rel_logits = rel_to_abs(rel_logits)
  # Shape it and tile height times
  rel_logits = tf.reshape(rel_logits, [-1, Nh, H, W, W])
  rel_logits = tf.expand_dims(rel_logits, axis=3)
  rel_logits = tf.tile(rel_logits, [1, 1, 1, H, 1, 1])
  # Reshape for adding to the logits.
  rel_logits = tf.transpose(rel_logits, transpose_mask)
  rel_logits = tf.reshape(rel_logits, [-1, Nh, H*W, H*W])
  return rel_logits
\end{lstlisting}
    \caption{Helper functions in Tensorflow for 2D relative self-attention.}
    \label{fig:helper_relative_code}
    \vspace{-1em}
\end{figure}

\lstset{
  backgroundcolor=\color{white},
  basicstyle=\fontsize{7.5pt}{8.5pt}\fontfamily{lmtt}\selectfont,
  columns=fullflexible,
  breaklines=true,
  captionpos=b,
  commentstyle=\fontsize{8pt}{9pt}\color{codegray},
  keywordstyle=\fontsize{8pt}{9pt}\color{codegreen},
  stringstyle=\fontsize{8pt}{9pt}\color{codeblue},
  frame=tb,
  otherkeywords = {self},
}
\begin{figure}[h!]
\tiny
\begin{lstlisting}[language=python]
def relative_logits(q, H, W, Nh, dkh):
  """Compute relative logits."""
  # Relative logits in width dimension first.
  rel_embeddings_w = tf.get_variable(
      'r_width', shape=(2*W - 1, dkh),
      initializer=tf.random_normal_initializer(dkh**-0.5))
  # [B, Nh, HW, HW]
  rel_logits_w = relative_logits_1d(
      q, rel_embeddings_w, H, W, Nh, [0, 1, 2, 4, 3, 5])

  # Relative logits in height dimension next. 
  # For ease, we 1) transpose height and width, 
  # 2) repeat the above steps and
  # 3) transpose to eventually put the logits
  # in their right positions.
  rel_embeddings_h = tf.get_variable(
      'r_height', shape=(2 * H - 1, dkh),
      initializer=tf.random_normal_initializer(dkh**-0.5))
  # [B, Nh, HW, HW]
  rel_logits_h = relative_logits_1d(
      tf.transpose(q, [0, 1, 3, 2, 4]),
      rel_embeddings_h, W, H, Nh, [0, 1, 4, 2, 5, 3])
  
  return rel_logits_h, rel_logits_w

def self_attention_2d(inputs, dk, dv, Nh, relative=True):
  """2d relative self-attention."""
  _, H, W, _ = shape_list(inputs)
  dkh = dk // Nh
  dvh = dv // Nh
  flatten_hw = lambda x, d: tf.reshape(x, [-1, Nh, H*W, d])

  # Compute q, k, v
  kqv = tf.layers.conv2d(inputs, 2 * dk + dv, 1)
  k, q, v = tf.split(kqv, [dk, dk, dv], axis=3)
  q *= dkh ** -0.5  # scaled dot-product

  # After splitting, shape is [B, Nh, H, W, dkh or dvh]
  q = split_heads_2d(q, Nh)
  k = split_heads_2d(k, Nh)
  v = split_heads_2d(v, Nh)

  # [B, Nh, HW, HW]
  logits = tf.matmul(flatten_hw(q, dkh), flatten_hw(k, dkh), transpose_b=True)

  if relative:
    rel_logits_h, rel_logits_w = relative_logits(q, H, W, Nh, dkh)
    logits += rel_logits_h
    logits += rel_logits_w

  weights = tf.nn.softmax(logits)
  attn_out = tf.matmul(weights, flatten_hw(v, dvh))
  attn_out = tf.reshape(attn_out, [-1, Nh, H, W, dvh])
  attn_out = combine_heads_2d(attn_out)
  # Project heads
  attn_out = tf.layers.conv2d(attn_out, dv, 1)
  return attn_out

def augmented_conv2d(X, Fout, k, dk, dv, Nh, relative):
  conv_out = tf.layers.conv2d(inputs=X, filters=Fout - dv, kernel_size=k, padding='same')
  attn_out = self_attention_2d(X, dk, dv, Nh, relative=relative)
  return tf.concat([conv_out, attn_out], axis=3)
\end{lstlisting}
    \caption{Tensorflow code for 2D relative self-attention.}
    \label{fig:relative_code}
    \vspace{-1em}
\end{figure}

\clearpage

\subsection{Attention visualizations.\label{sec:attn_viz}}

\begin{figure}[h!]
    \centering
    \includegraphics[width=\columnwidth]{{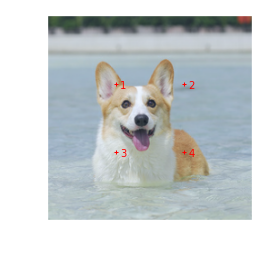}}
    \caption{An input image. The red crosses indexed 1 to 4 represent the pixel locations for which we show the attention maps in Figure ~\ref{fig:visualization}.}
    \label{fig:input_image}
\end{figure}

\begin{figure}[h!]
    \centering
    \includegraphics[width=\columnwidth]{{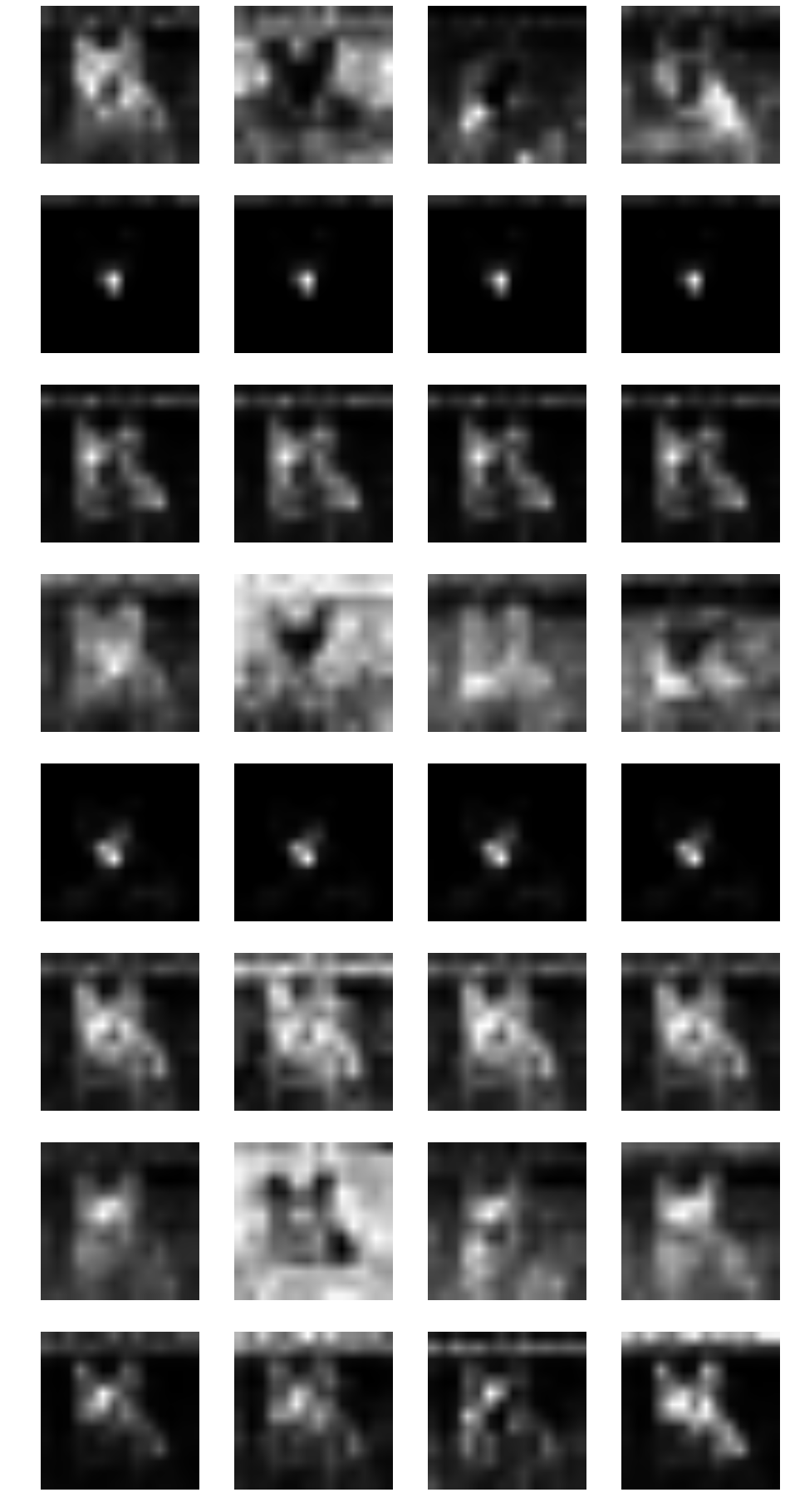}}
    \caption{Visualization of attention maps for an augmented convolution in the Attention-Augmented-ResNet50. Rows correspond to the 8 different heads and columns correspond to the 4 pixel locations depicted in the input image (See Figure ~\ref{fig:input_image}).}\label{fig:visualization}
\end{figure}

In Figure~\ref{fig:visualization}, we present attention maps visualizations for the input image shown in Figure~\ref{fig:input_image}. We see that attention heads learn to specialize to different content and notably can delineate object boundaries.

